\documentclass[letterpaper,twocolumn,10pt]{article}
\usepackage{usenix,epsfig,endnotes}

\usepackage{url}
\usepackage{bigfoot}
\usepackage{amsmath}
\usepackage{caption}
\usepackage{balance}
\usepackage{graphicx}
\usepackage{multirow}
\usepackage{xcolor}
\usepackage{algorithm}
\usepackage{algpseudocode}
\usepackage{tabularx}
\usepackage{amsfonts}
\usepackage[nocompress]{cite}

\newcommand{\imggroup}[3]{
 #2 &
\fbox{
\includegraphics[scale=0.38]{pair/image#1}
}
&
\fbox{
\includegraphics[scale=0.38]{pair/adv_image#1}
}
&
#3 
& \fbox{\includegraphics[scale=0.38]{pair/diff_image#1}}
\\
}
\newcommand{\paragraphbe}[1]{\vspace{0.75ex}\noindent{\bf \em #1}\hspace*{.3em}}

\newcommand{\X}{\mathbf{X}}
\newcommand{\Y}{\mathbf{Y}}
\newcommand{\A}{\Gamma}

\newcommand*{\vsepfbox}[1]{%
  \begingroup
    \sbox0{\fbox{#1}}%
    \setlength{\fboxrule}{0pt}%
    \mbox{\kern-\fboxsep\fbox{\unhbox0}\kern-\fboxsep}%
  \endgroup
}

\newcommand{\bnm}{\begin{newmath}}
\newcommand{\enm}{\end{newmath}}

\newcommand{\bea}{\begin{eqnarray*}}%
\newcommand{\eea}{\end{eqnarray*}}%

\newcommand{\bne}{\begin{newequation}}
\newcommand{\ene}{\end{newequation}}

\newcommand{\bal}{\begin{newalign}}
\newcommand{\eal}{\end{newalign}}

\newenvironment{newalign}{\begin{align*}%
\setlength{\abovedisplayskip}{4pt}%
\setlength{\belowdisplayskip}{4pt}%
\setlength{\abovedisplayshortskip}{6pt}%
\setlength{\belowdisplayshortskip}{6pt} }{\end{align*}}

\newenvironment{newmath}{\begin{displaymath}%
\setlength{\abovedisplayskip}{4pt}%
\setlength{\belowdisplayskip}{4pt}%
\setlength{\abovedisplayshortskip}{6pt}%
\setlength{\belowdisplayshortskip}{6pt} }{\end{displaymath}}

\newenvironment{newequation}{\begin{equation}%
\setlength{\abovedisplayskip}{4pt}%
\setlength{\belowdisplayskip}{4pt}%
\setlength{\abovedisplayshortskip}{6pt}%
\setlength{\belowdisplayshortskip}{6pt} }{\end{equation}}

\begin{document}

\date{}


\title{Fooling OCR Systems with Adversarial Text Images}

\author{
{\rm Congzheng Song}\\
Cornell University
\and
{\rm Vitaly Shmatikov}\\
Cornell Tech
} 

\maketitle

\thispagestyle{empty}

\subsection*{Abstract}

We demonstrate that state-of-the-art optical character recognition
(OCR) based on deep learning is vulnerable to adversarial images.
Minor modifications to images of printed text, which do not change
the meaning of the text to a human reader, cause the OCR system to
``recognize'' a different text where certain words chosen by the
adversary are replaced by their semantic opposites.  This completely
changes the meaning of the output produced by the OCR system and by the
NLP applications that use OCR for preprocessing their inputs.

\section{Introduction}

Machine learning (ML) techniques based on deep neural networks have led to
major advances in the state of the art for many image analysis tasks,
including object classification~\cite{krizhevsky2012imagenet}
and face recognition~\cite{taigman2014deepface}.
Modern ML models, however, are vulnerable to adversarial
examples~\cite{szegedy2013intriguing}: a minor modification to
an input image, often imperceptible to a human, can change the
output of an ML model applied to this image, e.g., produce an incorrect
classification~\cite{szegedy2013intriguing,papernot2016limitations,goodfellow2014explaining,carlini2017towards}
or cause the model to segment the image
incorrectly~\cite{cisse2017houdini}.

Optical character recognition (OCR) is another image analysis task where
deep learning has led to great improvements in the quality of ML models.
It is different from image classification in several essential ways.

First, modern OCR models are not based on classifying individual
characters.  Instead, they assign sequences of discrete labels
(corresponding to entire words) to variable-sized inputs.  Consequently,
they recognize text line by line, as opposed to character by character.
This presents a challenge for the adversary because, as we show, attacks
that simply paste adversarial images of individual characters into an
input image are ineffective against the state-of-the-art models.

\begin{figure}[t]
\centering
\begin{tabular}{ccc}
Input image & & OCR output \\ 
\parbox{0.22\textwidth}{
\includegraphics[width=0.21\textwidth]{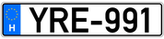}} & $\Rightarrow$ & \textbf{YRE-991} \\ 
$\downarrow$ add perturbation & &  \\
\parbox{0.22\textwidth}{
\includegraphics[width=0.21\textwidth]{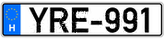}}  & $\Rightarrow$ & \textbf{YPF-881}
\end{tabular}
\caption{An adversarial example for an OCR-based license plate recognition
system.  The OCR model takes a variable-sized image as input and outputs
a sequence of characters that occur in the image.  This task is different
from image classification.}
\label{fig:license}
\end{figure}

Second, many applications of OCR involve recognizing natural-language
text (e.g., contents of a scanned document) and not just arbitrary
sequences of characters (as in the example of Fig.~\ref{fig:license}).
In this context, small perturbations to the input image typically cause
the OCR model to reject the input or else produce meaningless output.
The search for adversarial examples should be guided by linguistic
information\textemdash in our case, pairs of words that are visually
similar yet semantically opposite.

Some aspects of the OCR task favor the adversary.  When the goal of OCR
is to recognize natural-language text, incorrectly recognizing even a
single world can have a big impact on the overall meaning of the text.
An adversary who can effect a very small targeted change in the model's
output\textemdash for example, replace a well-chosen word with its
antonym\textemdash can completely change how a human would understand
the resulting text.

Third, OCR systems are often used as components of natural language
processing (NLP) pipelines.  Their output is fed into NLP applications
such as document categorization and summarization.  This amplifies the impact
of adversarial examples because NLP applications are highly sensitive to
certain words in their input.  As we show, tiny changes to input images
can dramatically change the output of the NLP models operating on the
results of OCR applied to these images.  Further, many NLP models are
trained on OCR-processed documents.  If some of these training inputs
are replaced by adversarial images, the adversary can poison the model.

\paragraphbe{Our contributions.}
We investigate the power of adversarial examples against
Tesseract~\cite{tesseract}, a popular OCR system based on deep learning.
We chose Tesseract because a trained model is publicly available (as
opposed to just the model architecture) and also because Tesseract is
used in many OCR-based applications.

We show how to generate adversarial images of individual words that cause
Tesseract to misrecognize them as their antonyms, effectively flipping
their meaning.  We then extend word-level attacks to entire documents.
Using the corpus of Hillary Clinton's emails for our experiments,
we show how to (1) modify key data, including dates, times, numbers,
and addresses, and (2) change a few chosen words to their antonyms,
completely changing the meaning of the text produced by OCR vs.\ the
meaning of the text in the original document.

We then evaluate our attack on NLP applications that rely on OCR to
extract text from images.  We show that adversarial text images can fool
a semantic analysis model into confidently producing wrong predictions.
For a document categorization model, the adversary can generate text
images that, after being processed by OCR, will be misclassified into
any target class chosen by the adversary.  We also show how adversarial
images can poison the training data and degrade the performance of a
sentiment analysis model.

The adversarial perturbations needed to stage successful attacks
against Tesseract (a) affect only a tiny fraction of the pixels in the
input image, and (b) are localized in a small subregion of the image,
corresponding to the few words being attacked.  Furthermore, (c) large
documents present many opportunities for an attack: the adversary has
many choices of words to modify in order to change the meaning of the
resulting text.

Limitations of our attack include transferability and physical
realizability.  In general, perturbations needed to make adversarial
text images physically realizable are so big that the resulting images
are rejected by Tesseract as too noisy.  Nevertheless, we demonstrate
an adversarial word image that fools Tesseract into recognizing a
semantically opposite word even after this image is printed onto paper
and scanned back into a digital form.

\section{Background}
\label{sec:background}
\begin{figure*}[t]
\centering
\includegraphics[width=0.8\textwidth]{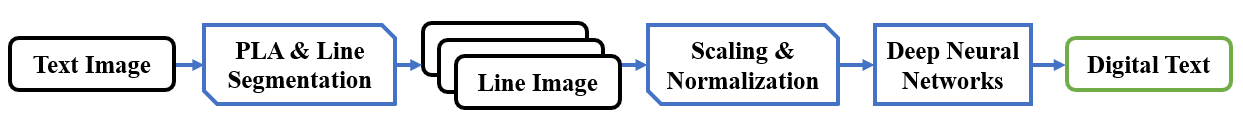}
\caption{OCR pipeline with a deep learning-based recognition
model.  The OCR system first performs page layout analysis (PLA) 
to detect the text in the image and
segments the image into sub-images containing one line of text each.
Each line image is scaled and normalized to match the training data
of the recognition model.  The normalized line images are fed into
the recognition model.  Finally, the OCR system outputs combined text
predictions.
}
\label{fig:ocrpipline}
\end{figure*}
\subsection{Deep learning for sequence labeling}

Deep learning has become very popular for many computer vision and image
recognition tasks~\cite{krizhevsky2012imagenet,taigman2014deepface}.
A deep learning model (or ``neural network'') is a function
$f_\theta:\X\mapsto\Y$ parametrized by $\theta$, where $\X$ is the input,
or feature, space and $\Y$ is the output space.  For classification
problems, $\X$ is a vector space (e.g., images of the same size) and $\Y$
is a discrete set of classes (e.g., the set of possible objects in the
images).  Supervised training of a model $f_\theta$ aims to find the best
set of parameters $\theta$ using the labeled training dataset $D = \{(x_i,
t_i)\}_{i}$ and the loss metric $L(f(x_i), t_i)$, which measures the
gap between the model's prediction $f(x_i)$ and the correct label $t_i$.

\emph{Sequence labeling} is a more complicated task that assigns a
sequence of discrete labels to variable-sized sequential input data.
For example, in optical character recognition, the input is an image
and the output is a sequence of characters $t = [t^1, t^2, \dots, t^N],
t^i\in\A$ from some alphabet $\A$.  Both the input image and the output
text can vary in length, and the alignment of image regions to the
corresponding text characters is not known a priori.

\textbf{Connectionist Temporal Classification}
(CTC)~\cite{graves2006connectionist} provides a alignment-free method
for training an end-to-end neural network for sequence labeling tasks.
In CTC, the neural-network model $f$ outputs a sequence of probability
vectors $f(x) = y = [y^1, y^2, \dots, y^M]$ where $M \geq N$ and $y^i\in
[0, 1]^{|\A|}$ is the probability distribution over all characters at
position $i$.

Training the model requires calculating the likelihood $p(t|x)$.  Because
$M$ is not necessarily the same as $N$, it is hard to directly measure
$p(t|x)$ from the model's prediction $f(x)$ and the target sequence $t$.
Instead, $p(t|x)$ is measured using a \textit{valid alignment} $a$ of $t$.
Sequence $a = [a^1, a^2,\dots, a^M]$ and $a^i\in \A \cup \{\text{blank}\}$
is a valid alignment of $t$ if $a$ can be turned into $t$ by removing
blanks and sequential duplicate characters.  For example, [c, a, a,
blank, t, t] is a valid alignment for [c, a, t].

Let $A$ be the set of all possible valid alignments of length $M$ for
the target sequence $t$.  Then the likelihood is
\[
p(t|x) = \sum_{a\in A}\prod_{i=1, M}p(a^i| x) = 
\sum_{a\in A}\prod_{i=1, M}(y^i)_{a^i} 
\]
The CTC loss function for the model prediction $f(x)$ and the target
sequence $t$ is
\begin{align}
\label{eq:ctcloss}
L_{\text{CTC}}(f(x), t) =  -\log p(t|x)
\end{align}
The training then proceeds as usual to minimize the CTC loss on all
training inputs.

The resulting model, given an input $x$, produces $f(x)$.  To obtain the
most probable output sequence, a greedy algorithm can select argmax $y^i$
at each position $i$ and collapse the alignment (i.e., eliminate blanks
and duplicates).  The greedy method, however, does not account for the
fact that a sequence can have many valid alignments.  A better method is
based on beam search decoding: keep a fixed number of the most probable
alignments at each position $i$ and return the (collapsed) output that
has the highest sum of probabilities for all valid alignments in the
top-alignment list.

\subsection{Optical character recognition}

Optical character recognition (OCR) is a technology that converts images
of handwritten or printed text (e.g., a scanned document, a page of
magazine, or even a photo of a scene that includes signs with text)
into digital text.

The OCR pipeline generally starts with preprocessing the images.
Common preprocessing techniques include page layout analysis for
localizing blocks of texts in the image, de-skewing the image if the
text is not aligned properly, and segmenting the image to extract blocks
or regions that contain text.  A recognition model is then applied to
the preprocessed images.  The characters produced by the model are the
output of the OCR pipeline.

Recognition models can be roughly categorized into two types:
character-based and end-to-end.

\textbf{Character-based recognition} is the traditional approach
to recognizing text in the ``block of text'' images.  Examples of
character-based OCR include GOCR~\cite{GOCR} and the legacy version
of Tesseract~\cite{smith2007overview}.  A character recognition model
first localizes characters in the image and segments the image into
sub-images that contain one character each.  The model then extracts
features from each sub-image and feeds them into a machine learning
classifier to identify the most likely character.  The features are
usually hand-engineered and may include, for example, lines, closed loops,
line directions and intersections, etc.  The classifier is typically a
fairly simple ML model such as K-nearest neighbors.

The performance of the classifier severely degrades if the
single-character images produced by the segmentation are bad.  Therefore,
the overall performance of character-based recognition models strongly
depends on the segmentation method.


\textbf{End-to-end recognition} is a segmentation-free technique
that aims to recognize entire sequences of characters in a
variable-sized ``block of text'' image.  Sequential models
such as Hidden Markov Models have been used for this
purpose~\cite{bengio1995lerec,lu1999advances,espana2011improving}.

With the recent advances in deep learning for image
analysis, end-to-end recognition models based on deep
neural networks~\cite{wang2012end,breuel2013high} have
become increasingly popular.  These models utilize neural
networks as the feature extractor and thus do not require that
features be manually engineered.  Sequential deep neural-network
models~\cite{graves2006connectionist,hochreiter1997long} also allow
variable-sized input images and thus avoid the issues that arise from
segmentation in character-based models.

\paragraphbe{OCR applications.} 
OCR is widely used for many real-world applications, including automated
data entry and license plate recognition~\cite{OpenALPR}.  OCR can also serve as the main
preprocessing step for natural language processing (NLP) tasks such as
text classification~\cite{abbyydoc,Yasser901992,Larsson934351}, document
retrieval~\cite{taghva1994results,kameshiro2001document,googlebook},
machine translation~\cite{googletranslate,yandextranslate}, and even
cancer classification~\cite{zuccon2012impact}.  All of these applications
critically depend on the correctness of OCR because the consequences
of mistakes are very serious\textemdash from wrong cars being fined for
violations to incorrect medical diagnoses.

\subsection{Tesseract}
\begin{table}[t]
\centering
\footnotesize
\begin{tabular}{l|c|c}
\hline
layer name & layer specs & layer output shape \\
\hline
\hline
conv2d & $ 3 \times 3 \times 16$ & $h \times w \times 16$ \\
\hline
maxpool2d & $3 \times 3$ & $h / 3 \times w / 3 \times 16$ \\
\hline
lstm-fys & 64  & $w / 3 \times 64$ \\
\hline
lstm-fx & 96 & $w / 3 \times 96$  \\
\hline
lstm-rx & 96 & $w / 3 \times 96$  \\
\hline 
lstm-fx & 512  & $w / 3 \times 512$  \\
\hline 
fc & 111 & $w / 3 \times 111$ \\
\hline 
\end{tabular}
\caption{Neural network architecture\protect\footnotemark of Tesseract's text recognition model.}
\label{tbl:tesseractnn}
\end{table}
\footnotetext{Details of the model specification are described
in \url{https://github.com/tesseract-ocr/tesseract/wiki/VGSLSpecs}}

We chose Tesseract for our investigation because it is one of the most
widely used open-source OCR systems~\cite{tesseract} and because a
trained model is available (as opposed to just the architecture).
The legacy version of Tesseract~\cite{smith2007overview} uses a
character-based recognition model, but we focus on the latest version
of Tesseract which uses an end-to-end deep learning-based recognition
model~\cite{tesseractnn,smith2016end}.  The OCR pipeline of the latest
version of Tesseract follows the flow chart in Fig.~\ref{fig:ocrpipline}.

Tesseract takes an image as input and performs page layout analysis to
find the regions that contains text.  Each region is then segmented into
images of individual lines of text.  These line images are then fed to
the deep learning model for text recognition.

Tesseract's recognition model takes a line image as input and outputs a
sequence of characters recognized in that line.  This line recognition
task is essentially a sequence labeling problem where the input images
can vary in width.  Tesseract adds a small preprocessing step that scales
and normalizes line images to match the input domain of Tesseract's
training data.

The overall architecture of Tesseract's deep learning model is given
in Table~\ref{tbl:tesseractnn}.  Inputs are gray-scaled images of size
$h\times w$.  The network starts with convolution layer (conv2d) with a
$3\times 3\times 16$ filter and $\tanh$ activation function, followed by
a $3\times 3$ max-pooling layer (maxpool2d).  The network is then stacked
with 4 long short-term memory~\cite{hochreiter1997long} (lstm) layers
with, respectively, 64, 96, 96, and 512 hidden units.  An LSTM layer can
have several modes
(shown in letters after the dash): f/r means forward/reverse pass,
x/y indicates if the direction of pass is horizontal or vertical,
s indicates whether it returns only the last step of LSTM outputs.
Finally, the output layer has 111 units, corresponding to the number
of possible English characters.  Therefore, the network produces $w /
3$ probability vectors of size $111$.  Given these vectors, Tesseract
outputs the most probable sequence of characters by beam-search decoding.

Tesseract's model has been trained on documents rendered from a
large-scale text corpus crawled from the Internet~\cite{tesseractdata}.
The parameters of the trained model are available
online~\cite{tesseractmodel}.

\subsection{Adversarial examples}

Many machine learning models are vulnerable to adversarial
examples~\cite{szegedy2013intriguing,goodfellow2014explaining,papernot2016limitations,carlini2017towards}.
For instance, in object classification tasks, a small
perturbation\textemdash perhaps even imperceptible to the human
eye\textemdash can cause the model to classify an image containing an
object of a certain type as a different type with high confidence.

More formally, given a model $f$ that maps an input $x$ to an output
prediction $t$, the adversary can perform either an untargeted attack, or
a targeted attack.  For the untargeted attack, the goal of the adversary
is to generate an adversarial example $x^\prime$ so that $f(x^\prime)
\neq t$.  For the targeted attack, the adversary has a specific target
output $t^\prime\neq t$ in mind.  His goal is to construct an adversarial
example $x^\prime$ so that $f(x^\prime) = t^\prime$.  For both types
of attacks, $D(x, x^\prime)$ must be below some threshold where $D$
is a distance metric that measures similarity between two inputs.

Constructing adversarial examples is usually formulated as an optimization
problem.  For a machine learning task, the loss function $L$ measures
the error between the true target $t$ and model's prediction $f(x)$.
The problem of generating an untargeted adversarial example for a valid
input $x$ with the distance threshold $\epsilon$ can be stated as:
\begin{align*}
\max_{x\prime} \quad &L(f(x^\prime), t)\\
\text{such that} \quad &D(x, x^\prime)\leq \epsilon
\end{align*}
Maximizing the loss term ``forces'' the model to make a wrong prediction
given $x^\prime$ while the distance to the valid input is below
$\epsilon$.  

For a targeted attack, the objective is $L(f(x^\prime), t^\prime)$,
which forces the model to predict $t^\prime$ instead of $t$.

The optimization problem is typically solved using standard gradient
descent.  Given access to the parameters of the model, the gradient with
respect to the adversarial input can be calculated using back-propagation.

Previous literature considered adversarial examples for standard image
classification tasks.  In Section~\ref{whyharder}, we explain why
generating adversarial examples for OCR models is more difficult.

\section{Attacking OCR Pipeline}

\subsection{Threat model}

We assume that the adversary has complete access to the entire OCR
pipeline, including the preprocessing algorithms, the architecture
and parameters of the recognition model, the decoding algorithm,
and\textemdash when the output of OCR is used as input into NLP
applications\textemdash the machine learning models used by the latter.

This assumption holds for Tesseract, as well as other open-source
OCR systems.  Prior work has also shown that it is sometimes
possible to generate adversarial examples in a black-box scenario
where the adversary does not know the core recognition model and its
parameters~\cite{papernot2016transferability}.  Other steps of the
pipeline, including preprocessing, are standardized in many systems and
thus easy for the adversary to reconstruct.

\subsection{Generating targeted adversarial examples for CTC-based OCR}
\label{sec:tessadv}

As described in Section~\ref{sec:background}, Tesseract (and other OCR
systems) performs sequence labeling.  Model $f$ takes image $x$ as input
and produces a sequence of characters $t$ as its output or ``prediction.''

An untargeted attack in this case will cause the model predict a
sequence of characters that does not match the ground truth (i.e.,
the characters that actually appear in the input image).  If, however,
the output of OCR is intended to be human-understandable natural text,
the untargeted attack may produce a sequence of gibberish characters
that does not form a valid text.

Instead, we focus on targeted attacks that cause the model to predict
\textbf{an adversary-specified sequence of characters $t^\prime$ which is
a valid text whose semantic meaning is different from the ground truth.}

Given an input image $x$, the ground truth sequence $t$, and
the target sequence $t^\prime$, we use the standard formulation
of the optimization problem to generate an adversarial example
$x^\prime$~\cite{szegedy2013intriguing}:
\begin{align*}
\min_{x^\prime}\quad & c\cdot L_\text{CTC}(f(x^\prime), t^\prime) +  || x - x^\prime ||^2_2\\
\text{such that}\quad & x^\prime \in [x_\text{min}, x_\text{max}]^p
\end{align*}
where $L_\text{CTC}$ is the CTC loss function for sequential labeling
detailed in Equation~\eqref{eq:ctcloss} and $|| x - x^\prime ||^2_2$
is the $L_2$-norm distance between the clean and perturbed images,
and the constant $c$ balances the two terms in the loss function.

The box constraint $x^\prime \in [x_\text{min}, x_\text{max}]^p$
where $p$ is the number of pixels ensures that the adversarial
example is a valid input for $f$.  In Tesseract, $x_\text{min},
x_\text{max}$ are $-1, 1$, respectively.  Using the change of variables
method~\cite{carlini2017towards}, we reformulate the minimization
problem as:
\begin{align*}
\min_\omega\quad & c\cdot L_\text{CTC}(f(\frac{\alpha\cdot\tanh(\omega) + \beta}{2}), t^\prime) \\ 
& + ||\frac{\alpha\cdot\tanh(\omega) + \beta}{2} - x||^2_2
\end{align*}
where $x^\prime = (\alpha\cdot\tanh(\omega) + \beta)/ 2$, $\alpha =
(x_\text{max} - x_\text{min}) / 2$ and $\beta = (x_\text{max} +
x_\text{min}) / 2$.  This formulation adds a new variable $\omega$ so
that $(\alpha\cdot\tanh(\omega) + \beta)/ 2$ satisfies the box constraint
automatically during optimization.

\paragraphbe{Differences between attacking CTC and attacking classification.}
\label{whyharder}
Generating targeted adversarial examples for the sequence labeling
models differs in several respects from attacking standard image
classification models, which were the subject of much prior
research~\cite{szegedy2013intriguing,carlini2017towards}.

First, the output of a CTC model is a varied-length sequence instead
of a single label.  A successful targeted attack thus needs to ensure
that the output sequence matches the target sequence exactly in terms of
length and each label in the sequence.  This is harder than attacking
the standard image classification task, which requires transforming a
single label produced by the model.

Second, a successful attack on a label in a given sequence may not work in
a different context.  For example, suppose we have an adversarial example
that causes \textit{include} to be misrecognized as \textit{exclude},
where we only changed letters \textit{in} to letters \textit{ex}.
The perturbation on the letters \text{in} may not work when applied
to other words, e.g., it may not cause \textit{internally} to be
misrecognized as \textit{externally}.  This is due to the nature
of the recurrent neural networks used in the CTC recognition model.
Their internal feature representations depend on the context and the
same perturbation may not transfer between contexts.

To the best of our knowledge, adversarial examples against sequence
labeling models have not been previously demonstrated in the image domain.
Concurrent work developed targeted attacks against speech-to-text
models~\cite{carlini2018audio}; we discuss the differences in
Section~\ref{sec:related}.

\subsection{Basic attack}

OCR is widely used for tasks such as processing scanned documents and
data entry, where the output of OCR is intended to be understandable
by humans.  The targeted attack from Section~\ref{sec:tessadv} can easily
modify data such as dates, addresses, numbers, etc., with serious impact
on documents such as invoices and contracts.

A more interesting attack takes advantage of the fact that OCR produces
natural-language text.  This attack transforms the meaning of the output
text by causing the OCR model to misrecognize a few key words.

\paragraphbe{Choosing target word pairs.} 
In languages such as English, there are pairs of words that are very far
in meaning but visually close enough that a small adversarial perturbation
is sufficient to fool the OCR system into recognizing an image of one
word as the other word.

One simple attack that can help transform the meaning of a text is to
replace a key word with its antonym.  To create a list of word pairs
for our experiments, we collected pairs of antonyms from the WordNet
dictionary~\cite{miller1995wordnet} where the distance between two
words in a pair is below a threshold.  In the experiments, we set the
threshold adaptively according to the number of characters in the word.
We also make sure the replaced word is the same part of speech as the
original word.  This ensures that the attack does not introduce (new)
grammatic errors into the text output by the OCR model.

\paragraphbe{Semantic filtering.} 
Although replacing a word with its antonym does not cause syntactic
errors, it may still cause semantic awkwardness in the transformed text.
There are several reasons for this.

First, an English word can have multiple meanings, thus simply replacing
a word with one of its antonyms many not fit the context.  For example,
changing \textit{``They carelessly fired the barn."} to \textit{``They
carelessly hired the barn."} turns the sentence into nonsense.
This issue can be potentially addressed using language modeling.
The adversary can check the linguistic likelihood of the transformed
text and, if it is very low, do not apply the attack.  In the above
example, the phrase \textit{hired the barn} should have a low score
because it is rare\textemdash although not entirely absent\textemdash
in English-language corpuses.

If the document contains multiple sentences, the replacement word
may not fit the context of the whole document even if its meaning is
indeed the opposite of the word it replaced.  The modified sentence
may not follow the logic of the surrounding sentences.  For example,
the attack can change \textit{``I am glad that \ldots''} to \textit{``I
am sad that \ldots''}.  However, if the context around this sentence
suggests a positive feeling, this replacement will look very awkward.
Changing the entire context may not be feasible if it requires careful
paraphrasing rather than simply replacing individual words.

Checking the semantic smoothness of a transformed document is a
non-trivial task.  As suggested in~\cite{jia2017adversarial}, it may
be possible to use crowd-sourcing to decide if the transformed document
makes sense or not.

\paragraphbe{Generating adversarial text images.} 
Given the original text of the document, first render a clean image.
Then find words in the text that appear in the list of antonym pairs
(see above).  Locate the lines of the clean image containing the words
to be transformed, transform them, and keep only the transformations
that produce valid words and pass semantic filtering (i.e., do not
produce semantic inconsistencies in the resulting text).  Then generate
adversarial examples for these line images and replace the images of the
corresponding lines in the document image.  The OCR model with recognize
all lines of the image correctly except for the modified lines.  For the
modified lines, the model will output the correct text with some of the
words replaced by their antonyms.

\subsection{Fooling NLP applications}

OCR systems are often used as just one component in a bigger
pipeline, which passes their output to applications operating on the
natural-language text (e.g., document categorization or summarization).
These pipelines are a perfect target for the adversarial-image attacks
because the output of OCR is not intended to be read or checked by
a human.  Therefore, the adversary does not need to worry about the
syntactic or semantic correctness of the OCR output as long as this
output has the desired effect on the NLP application that operates on it.

\paragraphbe{Generating adversarial text for NLP models.} 
We provide a simple greedy algorithm (Algorithm~\ref{alg:foolnlp}) to
automatically generate, given the original text $t$, the target text
$t^\star$ that we want the OCR system to produce as output.  This target
text will serve as input to an NLP classifier $h$ so that the class
predicted by $h(t^\star)$ will be different from the correct prediction
$h(t)$.  For simplicity, we assume that $h$ is a binary classifier where
$h(t)$ is the score for how likely the input is to be classified as the
correct class, e.g., the probability in a logistic regression model or
distance to the hyperplane in an SVM.

We first find the optimal replacement for each word $w$ in $t$.  We select
the candidate set of replacement words $W$ so that the edit distance
between $w$ and each word in $W$ is below some threshold $\tau$.  The
restriction on the edit distance allows us to use smaller perturbations
when generating adversarial text images.  We then compute the scores
on the modified texts where $w$ is replaced by each $w\prime\in W$.
We select $w^\prime$ that biases the original score $h(t)$ the most as
the optimal replacement for $w$.

We then sort all optimal word replacements in descending order by
the changes they cause in the score.  The goal is to identify words
that are most influential in changing the prediction of the NLP model.
We then greedily modify $t$ to $t^\star$, replacing the most influential
words by their optimal replacements, and repeat the procedure until
$h(t^\star)$ meets some model failure criterion (e.g., the score is
below some threshold or the model prediction is wrong).

This approach easily generalizes to multi-class models, whose output
$h(t)$ is a $k$-dimensional vector where $k$ is the number of classes.
If we want the model to incorrectly predict class $i$ as $j$, we modify
Algorithm~\ref{alg:foolnlp} to select word replacements that maximize
$\delta=h(t^\prime)_j - h(t)_j$ and keep the rest of the algorithm
unchanged.

\begin{algorithm}[t]
\caption{Generating target text for NLP classifier}
\begin{algorithmic}[1]
\State \textbf{Input:} NLP classifier $h$, victim text $t$, vocabulary
$V$, edit-distance threshold $\tau$, model failure criterion
\State \textbf{Output:} Modified target text $t^\star$ or failure $\perp$
\State $R\gets\emptyset$ \Comment{\textit{set of word replacements}}
\For{each word $w$ in $x$}
\State $W\gets\{w^\prime | w^\prime \in V, \textrm{edit-distance}(w, w^\prime) \leq \tau \}$ 
\State $T\gets\{t^\prime | t \text{ with } w \text{ replaced by }w^\prime, w^\prime \in W \}$ 
\State $\Delta\gets\{\delta| \delta = h(t^\prime) - h(t), t^\prime \in T\}$
\If{$\min\Delta < 0$}
\State $w^\prime \gets $ word replaced in $\text{argmin}_t\Delta$
\State $\delta\gets\min\Delta$
\State $R\gets R\cup \{(w, w^\prime, \delta)\}$
\EndIf
\EndFor
\State Sort $R$ by absolute value of $\delta$ in descending order.
\State $t^\star\gets t$
\For{$(w, w^\prime, \delta)\in R$}
\State Replace $w$ in $t^\star$ with $w^\prime$
\If{$h(t^\star)$ meets model failure criterion}
\State\Return{$t^\star$}
\EndIf
\EndFor
\State\Return{$\perp$}
\end{algorithmic}
\label{alg:foolnlp}
\end{algorithm}

\paragraphbe{Generating adversarial text images.} 
We first render a clean image based on the original text $t$. We run
Algorithm~\ref{alg:foolnlp} on $t$ to obtain the adversarial text
$t^\star$. We then locate the lines of the clean image where the
texts needs to be modified.  As in the basic attack, we generate the
adversarial examples for these lines and replace the original lines with
the generated images.

\paragraphbe{Data poisoning attacks.} 
\label{poison}
OCR is often used as a preprocessing step for collecting raw data to
train NLP models~\cite{abbyydoc,taghva1994results}.  Our attack can
contaminate the raw text images used as part of the training data and
thus affect the performance of the trained model.

We assume that the adversary has access to the raw text images and
can modify a subset of these images.  He first trains a benign NLP
model $h_0$ based on the OCR output on clean images.  He then uses
Algorithm~\ref{alg:foolnlp} to generate adversarial texts for a subset of
the training texts based on $h_0$.  He then generates adversarial images
accordingly and uses them to replace the clean images.  The adversarial
images look benign but the texts extracted from them by the OCR model
are different from the original texts.  The final NLP model will be
trained on the adversarial texts and thus its performance will degrade.

\section{Experiments}
\begin{figure*}[t]
\centering
\footnotesize
\begin{tabular}{lllll}
\hline
Ground truth & Clean image & Adversarial image & OCR output & Perturbation \\ 
\hline\hline
\imggroup{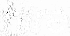}{hire}{fire}
\imggroup{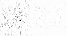}{glad}{sad} 
\imggroup{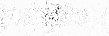}{waning}{waxing}
\imggroup{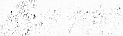}{dissent}{assent}
\imggroup{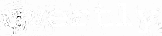}{overtly}{covertly}
\imggroup{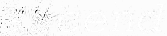}{ascend}{descend}
\imggroup{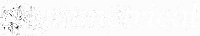}{asymmetrical}{symmetrical} 
\imggroup{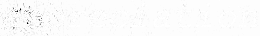}{appreciation}{depreciation} 
\hline
\end{tabular}
\caption{Adversarial renderings of words misrecognized as their antonyms
by Tesseract.  Perturbation is the absolute difference between the clean
and adversarial images.}
\label{fig:word}
\end{figure*}

\subsection{Setup}

We used the latest Tesseract version 4.00 alpha, which employs the
deep learning model described in Table~\ref{tbl:tesseractnn} for
recognition.  We downloaded the parameters of Tesseract's recognition
model and loaded them into our Tensorflow~\cite{abadi2016tensorflow}
implementation of the same recognition model.\footnote{The
implementation is copied from the scripts available at
\url{https://github.com/tensorflow/models/tree/master/research/street}
with minor modifications to take advantage of GPUs.} We implemented
the attack described in Section~\ref{sec:tessadv} with the Adam
optimizer~\cite{kingma2014adam}, generated adversarial examples using
our Tensorflow implementation, and evaluated them by directly applying
Tesseract.

\subsection{Attacking single words}
\begin{table}[t]
\centering
\footnotesize
\begin{tabular}{l|r|r|r|r}
\hline
Font &  Clean acc & Target acc & Rejected & Avg $L_2$ \\
\hline\hline
Arial & 100.00\% & 94.17\% & 0.00\% & 3.10 \\
Arial B & 100.00\% & 96.67\% & 0.00\% & 3.27 \\
Arial BI & 100.00\% & 95.00\% & 0.00\% & 3.14 \\
Arial I & 99.17\% & 94.17\% & 0.83\% & 2.90 \\
Courier & 99.17\% & 79.17\% & 0.00\% & 2.73 \\
Courier B & 100.00\% & 96.67\% & 0.00\% & 3.36 \\
Courier BI & 100.00\% & 93.33\% & 0.00\% & 3.23 \\
Courier I & 99.17\% & 93.33\% & 0.83\% & 2.78 \\
Georgia & 100.00\% & 91.67\% & 0.83\% & 2.94 \\
Georgia B & 100.00\% & 94.17\% & 0.83\% & 3.18 \\
Georgia BI & 100.00\% & 92.50\% & 0.83\% & 3.03 \\
Georgia I & 100.00\% & 95.00\% & 0.00\% & 2.99 \\
Times NR & 100.00\% & 88.33\% & 0.00\% & 2.90 \\
Times NR B & 100.00\% & 91.67\% & 0.00\% & 3.04 \\
Times NR BI & 98.33\% & 96.67\% & 0.00\% & 2.81 \\
Times NR I & 96.67\% & 90.00\% & 0.00\% & 2.75 \\
\hline
\end{tabular}
\caption{Results of attacking single words rendered with different fonts
(B indicates bold, I indicates italic). Clean acc is the accuracy of
Tesseract (percentage of predictions that match the ground truth) on
clean images.  Target accuracy is the accuracy of Tesseract predicting
the target word (antonym) on adversarial images.  Rejected is the
percentage of adversarial images that are rejected by Tesseract due
to large perturbation. Avg $L_2$ is the average $L_2$ distance between
clean and adversarial images.}
\label{tbl:word}
\end{table}

We selected 120 pairs of antonyms from WordNet~\cite{miller1995wordnet}
that meet our threshold requirement on the edit distance.  We set the
threshold adaptively according to the number of characters in the word
(2, 3, or 4 if the number of characters is, respectively, 5 or less,
6 to 9, or above 9).  Some examples of the pairs in our list are
\textit{presence/absence, superiority/inferiority, disabling/enabling,
defense/offense}, and \textit{ascend/descend}.

We render these words with 16 common fonts and set their antonyms as the
target output.  We set the number of iterations of gradient descent to
1,000, step size for optimization to 0.01, and the constant $c$ in the
objective function to 20.  Some of the resulting adversarial images are
shown in Fig.~\ref{fig:word}.  The perturbation is very minor but the
output of Tesseract is the opposite of the word appearing in the image.

The overall results for the word-pairs attack are summarized in
Table~\ref{tbl:word}.  The performance of the attack varies for
different fonts, but for most fonts we can successfully cause over
90\% of the words in our list to be misrecognized as their antonyms.
The amount of perturbation as measured by the $L_2$ distance is similar
for different fonts.  If too much perturbation is applied to an image,
Tesseract rejects the input and does not output anything.

\subsection{Attacking whole documents}

We now illustrate how our attack works for the images of whole documents,
using documents from the publicly available corpus of Hillary Clinton's
emails\footnote{Hillary Clinton's emails corpus is available at
\url{https://www.kaggle.com/kaggle/hillary-clinton-emails/data}}.

\paragraphbe{Changing key data in text.} 
We first show how our attack can be used to change key data (e.g., names,
numbers and addresses) in a document.  We chose a document from the
corpus that contains such information and rendered it as a clean image.
We then set the target output according to the type of information (name
to a different name, date to a different date etc).  Some care must
be taken when choosing the target values in order to preserve semantic
consistency.  For example, if the ground-truth text is \textit{Tuesday,
May 19}, then the target \textit{Thursday, May 18} is not semantically
valid because the day of the week and the actual date do not match.

\begin{figure*}[t!]
\centering
\vsepfbox{\parbox{\textwidth}{
{\footnotesize\textbf{Input image:}}\\
\vsepfbox{
\includegraphics[width=0.975\textwidth]{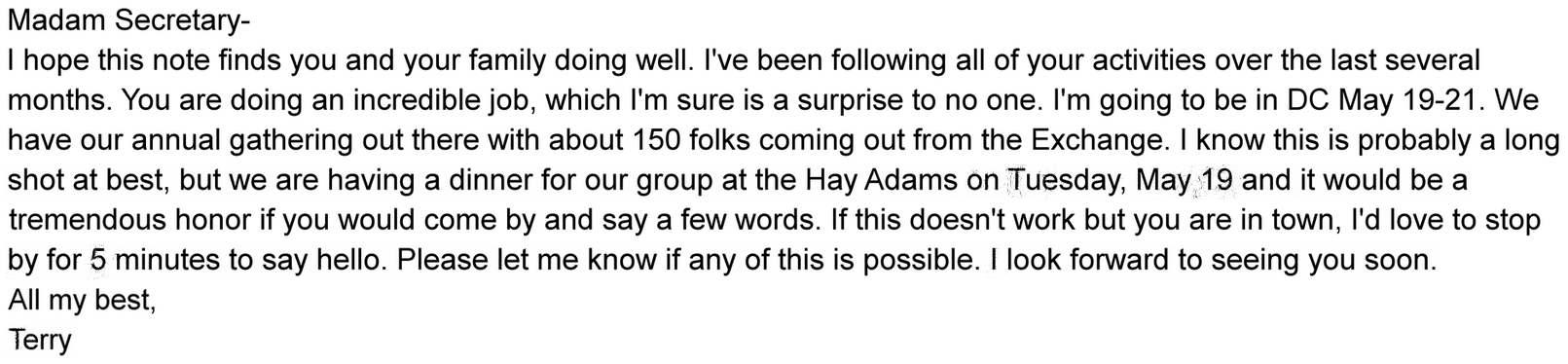}
}
\\ 
\footnotesize
\textbf{Tesseract output:}
Madam Secretary-
I hope this note finds you and your family doing well. I've been following all of your activities over the last several
months. You are doing an incredible job, which I'm sure is a surprise to no one. I'm going to be in DC May 19-21. We
have our annual gathering out there with about 150 folks coming out from the Exchange. I know this is probably a long
shot at best, but we are having a dinner for our group at the Hay Adams on {\color{red} Thursday, May 21} and it would be a
tremendous honor if you would come by and say a few words. If this doesn't work but you are in town, I'd love to stop
by for {\color{red} 50} minutes to say hello. Please let me know if any of this is possible. I look forward to seeing you soon. All my best, {\color{red} Jerry}
}}
\vspace{1em}
\centerline{(a) Example of changing information such as date, numbers and name.}

\vsepfbox{\parbox{\textwidth}{
{\footnotesize\textbf{Input image:}}\\
\vsepfbox{
\includegraphics[width=0.975\textwidth]{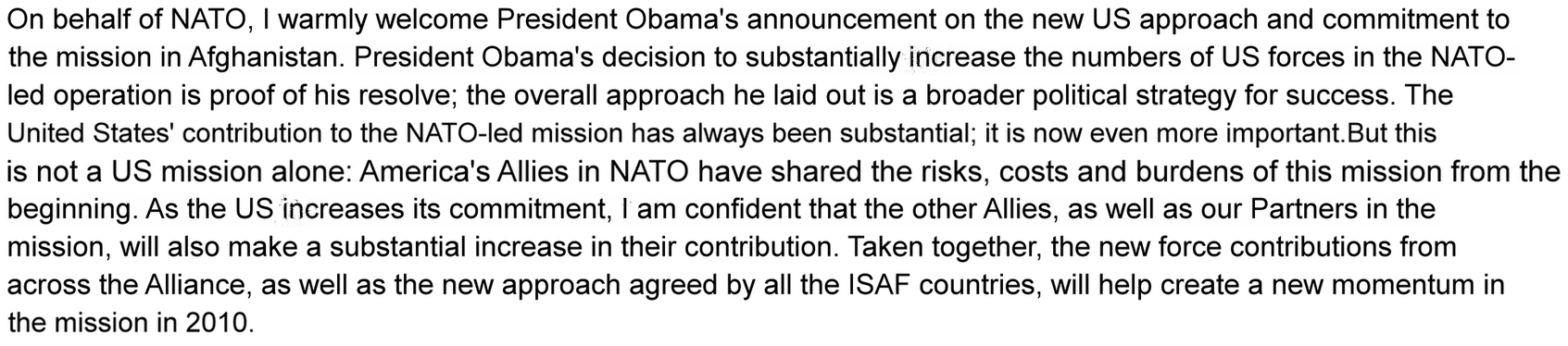}
}
\\ 
\footnotesize
\textbf{Tesseract output:}
On behalf of NATO, I warmly welcome President Obama's announcement on the new US approach and commitment to
the mission in Afghanistan. President Obama's decision to substantially {\color{red} decrease} the numbers of US forces in the NATO-
led operation is proof of his resolve; the overall approach he laid out is a broader political strategy for success. The
United States' contribution to the NATO-led mission has always been substantial; it is now even more important.But this
is not a US mission alone: America's Allies in NATO have shared the risks, costs and burdens of this mission from the
beginning. As the US {\color{red} decreases} its commitment, I am confident that the other Allies, as well as our Partners in the
mission, will also make a substantial increase in their contribution. Taken together, the new force contributions from
across the Alliance, as well as the new approach agreed by all the ISAF countries, will help create a new momentum in
the mission in 2010.
}}
\centerline{(b) Example of changing semantic meaning of the text.}
\caption{Document-level attack on one of Hillary Clinton's emails.
Texts in red are the adversary-chosen targets that Tesseract outputs
even though the text in the image is different.}
\label{fig:changenums}
\end{figure*}

Fig.~\ref{fig:changenums} (a) shows an example of a successful attack,
which changes the recognized date, time, and name information with a
small perturbation on the document image.  This shows a potential risk for
OCR systems used for data entry from scanned images, where the output of
OCR is not structured natural language but discrete pieces of information.

\paragraphbe{Changing semantic meaning of text.} 
We can also change the meaning of a document using the antonym pairs
from the word-level attacks.  Although our antonym list is very short,
these words are frequently used.  For example, in Hillary Clinton's email
corpus, words from our antonym list occur in 2,207 out of 7,945 emails,
with each email contains on average 3.05 words from our list.

We show an example of a successful attack in Fig.~\ref{fig:changenums}
(b), where the text output by Tesseract conveys a meaning opposite to
the original email.  The original email expresses the idea that the
U.S. will increase its forces in a NATO-led operation, and so will its
allies.  We render an adversarial text image so as to cause Tesseract
to output \textit{increase} instead of \textit{decrease} in two key
positions.  The text ``recognized'' by Tesseract now expresses the idea
that U.S. will decrease the commitment while the allies will increase
theirs, which is the exact opposite of the meaning of the original email.
This illustrates how the meaning of a relatively long document can be
flipped with a well-chosen change to one or two words.

\begin{figure*}
\centering
\begin{tabular}{cc}
\includegraphics[width=0.35\textwidth]{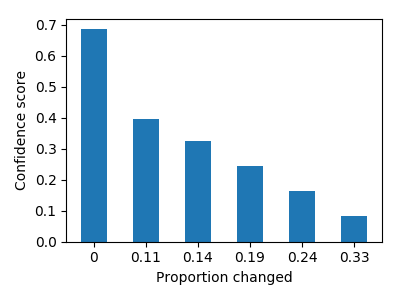} \hspace*{4em}&
\includegraphics[width=0.35\textwidth]{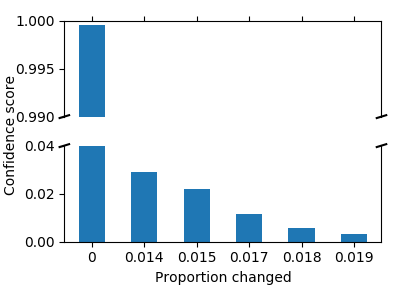}\\
(a) Rotten Tomatoes \hspace*{4em}& (b) IMDB
\end{tabular}
\caption{Average proportion of changed words in each text vs.\ the average
confidence score for the original prediction.}
\label{fig:confidence}
\end{figure*}

\subsection{Attacking NLP applications}

We now demonstrate that our attack can significantly affect NLP
applications if they operate on the results of OCR applied to text images.

\paragraphbe{Sentiment analysis.} 
Sentiment analysis is a binary classification task to determine
whether the input text contains positive or negative sentiment.
We chose Rotten Tomatoes (RT)~\cite{Pang+Lee:04a} and IMDB movie review
datasets~\cite{maas-EtAl:2011:ACL-HLT2011} for evaluation.  For the
RT dataset, we train a logistic regression classifier with bag-of-word
features. For the IMDB dataset, we train a convolutional neural network
(CNN) with word embedding features~\cite{kim2014convolutional}.  Given an
input text, both models output a confidence score for the polarity of
sentiment in the input text. The logistic regression model achieves
78.4\% accuracy on RT's test data and the CNN model achieves 90.1\%
accuracy on IMDB's test data.

We use Algorithm~\ref{alg:foolnlp} to generate adversarial text.
For each dataset, we construct a list of valid replacements for each
word in the vocabulary by setting the edit-distance threshold to 2.
First, we show how to control the polarity and confidence of sentiment
prediction by varying the number of words replaced in a text.

We chose the first 1,000 correctly classified texts in the test datasets
for both RT and IMDB as the targets for our attack.  We set the model
failure criterion to be the confidence score lower than 0.1 to 0.5 for
RT and 0.01 to 0.05 for IMDB.  The average proportion of words we need
to replace in a text and the corresponding confidence score of the model
prediction are shown in Fig.~\ref{fig:confidence}.  For the RT dataset,
we can change the score for the original label to a value below 0.1
(equivalently, over 0.9 for the opposite label) by replacing 30\% of
the text on average.  For the IMDB dataset, many fewer words need to
be replaced: by changing only 1\% to 2\% of the text, we can cause the
score for the original label to drop below 0.01.

\begin{figure}
\centering
\fbox{\parbox{0.47\textwidth}{
\footnotesize \textbf{Input image:}\\
\vsepfbox{
\includegraphics[width=0.445\textwidth]{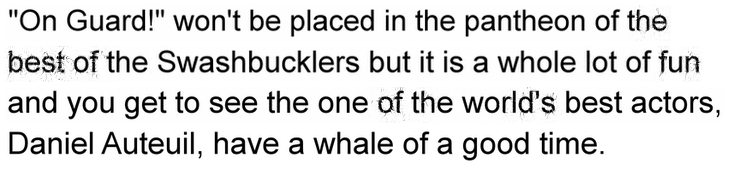}
}
\\
{\footnotesize \textbf{Tesseract output:}
``On Guard!" won't be placed in the pantheon of {\color{red} too mess too} the Swashbucklers but it is a whole lot of {\color{red} junk} and you get to see the one {\color{red} too} the {\color{red} worst} best actors, Daniel Auteuil, have a whale of a good time.}
}}
\caption{Adversarial rendering of a movie review from Rotten Tomatoes. The
sentiment analysis model predicts a positive score of \textbf{0.91}
on the original text but only \textbf{0.07} on Tesseract's output for
the adversarial image of the same review.}
\label{fig:sentiment}
\end{figure}

We now evaluate whether we can successfully carry out this attack in
the image domain.  We select 250 of the successfully attacked examples
for both datasets and render clean images based on the original texts.
We then generate adversarial examples for these texts.  We set the
balance constant $c$ to 25 and the number of iterations to 1,000.
For RT, our adversarial images achieve 92\% target accuracy.  As a
result, the accuracy of the sentiment classifier drops from 100\%
to 5\% when applied to the OCR output on these adversarial images
(see Fig.~\ref{fig:sentiment} for a successful example). The average
$L_2$ distance for RT's adversarial examples is 3.04 per changed word.
For IMDB, our adversarial images achieve 88.7\% target accuracy and
the sentiment classifier's accuracy drops from 100\% to 0\% on the
OCR-recognized text.  Note that the adversarial images do not need to be
perfectly recognized as the targets set by the adversary, as long the
OCR output fools the sentiment classifier.  The average $L_2$ distance
for IMDB's adversarial examples is 3.25 per changed word.

\paragraphbe{Document categorization.} 
We now show that our attack can generalize to a multi-class document
categorization task.  We evaluate our attack on the 20 Newsgroup
dataset\footnotemark, where the task is to categorize news documents into topics.
The original dataset contains 20 classes, but we select a subset of 4
classes for our experiment.  We train a one-vs-all logistic regression
classifier on the bag-of-word features as the target model to attack.
This model achieves 84.7\% accuracy on the test dataset.
\footnotetext{20 Newsgroup dataset is available at \url{http://qwone.com/~jason/20Newsgroups/}.}

We chose the first correctly classified 500 texts in the test dataset
to attack.  Similar to the sentiment analysis experiment, we built a
list of valid word replacement for each word in the vocabulary with
edit distance below 2.  For each text, we generated 3 adversarial
texts that try to change the model's output from the original class
to one of the three other classes.  The results of this attack are
shown in Table~\ref{tbl:multiclass}.  The percentage of modified texts
(mis)classified to the adversary-chosen target is very high for all
source classes and target classes.  The proportion of words that need
to be changed depends on the original and target classes.  For example,
transforming class 2 (religion) to other classes requires changing more
words than our transformations.

We next select 250 of the successfully attacked examples for 
generating adversarial images.
We render clean images based on the original texts and generate the
adversarial examples, setting the balance constant $c$ to 30 and the
number of iterations to 2,000.  The resulting adversarial images cause
Tesseract to output the desired adversarial texts with 84.8\% accuracy. 
$86.3\%$ of these texts are classified to the adversary-specified target
class.  The average $L_2$ distance for these adversarial examples is
2.56 per changed word.

\begin{table}
\footnotesize
\centering
\begin{tabular}{c|c|c|c|c}
\hline
class & 1 & 2 & 3 & 4 \\ \hline\hline
1 & - & 100 / 6.55 &  100 / 6.58 & 100 / 7.01 \\
2 & 98.1 / 25.7  & - & 100 / 12.6  & 97.5 / 33.9 \\ 
3 & 99.4 / 9.57 & 100 / 6.17 & - &  100 / 13.8 \\
4 & 100 / 6.39 & 100 / 5.64 & 100.0 / 5.33  & - \\ 
\hline
\end{tabular}
\caption{Class transformation accuracy on the 20 Newsgroup dataset.
Classes 1 through 4 are \textit{atheism, religion, graphics, space}
respectively.  An $a/b$ entry in row $i$ and column $j$ of the table
means that, on average, fraction $b$ of the words in each text needs to
be changed so that texts from class $i$ are misclassified by the model
as class $j$ with accuracy $a$.}
\label{tbl:multiclass}
\end{table}

\paragraphbe{Data poisoning.} 
As described in Section~\ref{poison}, if an NLP model is trained on
OCR-processed images, adversarial images can poison the training dataset.
We evaluate this attack on the Rotten Tomatoes dataset with a logistic
regression classifier.

\begin{figure}
\centering
\includegraphics[width=0.35\textwidth]{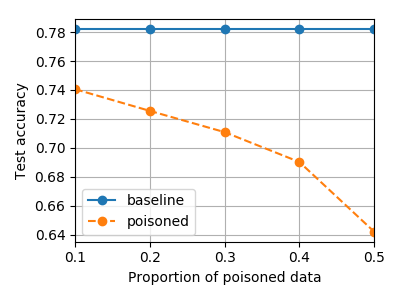}
\caption{Proportion of poisoned training texts vs.\ classification
accuracy on test data.  Baseline (solid) is the performance of the model
trained on the clean training data and poisoned (dashed) is the performance
of the model trained on the contaminated training data.}
\label{fig:poison}
\end{figure}

We first train a benign model $h_0$ on the original training data. We
then generate adversarial versions for a subset of the training texts.
We set the model failure criterion to confidence score below 0.1
and generate texts that will be confidently misclassified by $h_0$.
We retrain the model on the poisoned dataset where the adversarial texts
are substituted for the corresponding original texts.  The results are
shown in Fig.~\ref{fig:poison} for different fractions of the training
data replaced by adversarial texts.

We select 300 of the adversarial texts as targets for adversarial image
rendering.  We set the balancing factor $c$ to 200 as the proportion
of words changed in each text is much larger (0.87 on average) than in
the previous experiments.  The number of iterations is set to 2,000.
Tesseract outputs the desired adversarial texts with 91.3\% accuracy.
The average $L_2$ distance for these adversarial examples is 2.06 per
changed word.

\section{Limitations}

\paragraphbe{Transferability across contexts.} 
Internal features used by recurrent neural networks, such as the network
at the core of Tesseract's OCR model, are context-dependent and vary
from input to input.  Therefore, how Tesseract recognizes a particular
word depends on the words surrounding it.

Consequently, an adversarial image of a word which is misrecognized
by Tesseract in a particular document cannot be simply pasted into an
image of another document.  Even for the same word, adversarial images
must be rendered separately for each document.  For the same reason
(context-dependent features), adversarial images of individual letters
do not transfer from word to word.

\paragraphbe{Transferability across OCR models.} 
For basic image classification, previous
work~\cite{liu2016delving,papernot2016transferability} demonstrated that
adversarial examples generated for one model can also fool other models.
This shows that, in principle, adversarial examples can work in a
black-box setting.

OCR systems are significantly more complex.  They employ multi-step image
processing, which destroys or modifies many features of the input images.
Achieving transferability is much harder in this setting.

For character-based OCR models, such as GOCR~\cite{GOCR} and the legacy
version of Tesseract, our adversarial examples do not transfer because
the input processing pipeline is very different from the end-to-end OCR
models, which are the focus of this paper.  In particular, they segment
their inputs into a sequence of character-level images, which are then
fed into a machine learning model that is \emph{not} based on a recurrent
neural network.

Our adversarial examples do not transfer to other end-to-end
OCR models, either, because they apply different preprocessing
to input images.  For example, OCRopus~\cite{OCRopus} also uses
recurrent neural networks as the core of its recognition model, but its
preprocessing\footnote{Details of preprocessing in OCRopus are described
in \url{https://github.com/tmbdev/ocropy/wiki/Page-Segmentation}} includes
binarizing each pixel in the image, which truncates 8-bit values to 1
bit and thus destroys almost the entire adversarial perturbation.

Furthermore, generating transferable adversarial examples for targeted
attacks is much harder than for untargeted attacks~\cite{liu2016delving}.
Whether it is possible to achieve transferability of targeted adversarial
images across OCR models is an important topic for future work.

\paragraphbe{Physical realizability.} 
\label{sec:physical}
Recent work demonstrated robust adversarial examples that can
be feasibly realized in the physical world and not just as digital
images~\cite{kurakin2016adversarial,sharif2016accessorize,evtimov2017robust,athalye2017synthesizing}.
These examples are generated by taking into account physical-world
conditions such as lighting, scaling, angle of view, etc.

\begin{figure}[t]
\centering
\fbox{
\includegraphics[width=0.25\textwidth]{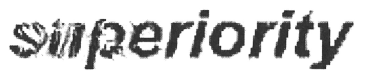}}
\caption{A physically realizable adversarial example. This example
is scanned from a printed image.  Tesseract misrecognizes it as
\textbf{inferiority}.}
\label{fig:scan}
\end{figure}

Similar techniques can help generate physically realizable adversarial
images that work against Tesseract.  We add different levels of scaling
transformations~\cite{athalye2017synthesizing} to the adversarial image
during optimization and also optimize the CTC loss to a very small value
so that the model is confident in predicting the target word.  We print
the resulting adversarial image on A4 paper, scan it back to digital
format with DPI set to 72, and submit the scanned image to Tesseract.
Fig.~\ref{fig:scan} shows a successful example.  The $L_2$ distance
for this example is 14.3, which is 5 times larger than the digital
adversarial example in Fig.~\ref{fig:word}.  Indeed, the image is much
noisier visually.

\begin{figure}
\centering
\includegraphics[width=0.4\textwidth]{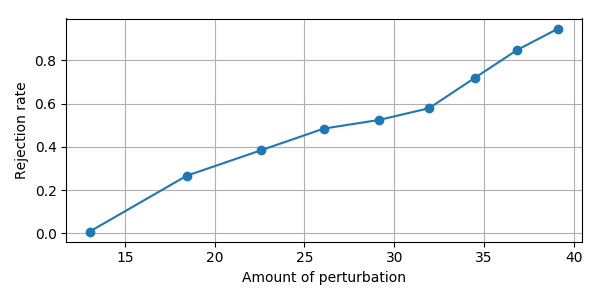}
\caption{Amount of perturbation (measured by $L_2$ distance) vs.\
rejection rate of Tesseract.}
\label{fig:rej}
\end{figure}

Tesseract rejects input images that it perceives as having low quality.
This presents a significant obstacle to physical realizability because
physical realizability requires large perturbations (to survive the
scanning) which make the scanned image too noisy for Tesseract.
Fig.~\ref{fig:rej} shows the relationship between the amount of
perturbation (measured by $L_2$ distance) and Tesseract's rejection
rate, calculated on 500 images of individual words.  As perturbations
become more significant, rejection rate increases.  This largely foils
the existing approaches to generating physical adversarial examples,
although, as Fig.~\ref{fig:scan} shows, some of our examples succeed.

\section{Mitigation}

A lot of recent
research~\cite{kurakin2016adversarial,papernot2016distillation,guo2017countering,tramer2017ensemble}
has aimed to increase the robustness of ML models to adversarial
inputs.  To the best of our knowledge, there is no universal technique
that is guaranteed to defend image analysis models against these
attacks~\cite{athalye2018}.

One common way to increase the robustness of ML models is through
adversarial training~\cite{kurakin2016adversarial,tramer2017ensemble},
where the training dataset is augmented with adversarial examples.
In contrast to the standard image classification tasks whose purpose
is to label images into a relatively small, finite number of classes or
to recognize the presence of a relatively small number of object types,
potential outputs of an OCR system include all possible character strings,
complicating the search for a comprehensive set of adversarial examples
to include in training.

Some assumptions about adversarial examples that are common in the image
classification literature may not hold for the OCR models.  For example,
it is often assumed that images that are close to each other must belong
to the same class~\cite{goodfellow2014explaining,madry2017towards}.
In the natural-language context, however, visual similarity between words
does not imply anything about their semantic proximity.  Images of the
same word may be very different, but two pixel-wise similar images may
depict words that are semantically far away from each other.  Furthermore,
OCR models such as Tesseract tolerate very little noise in their inputs
(see Section~\ref{sec:physical}).  Therefore, a small perturbation of the
original image may cause the model to reject it or else \emph{correctly}
output a different sequence.

Adversarial examples investigated in this paper change individual
words and do not automatically guarantee the semantic consistency of
the output as a whole.  After the adversary replaces certain words,
the resulting text may appear unnatural or logically inconsistent.
In theory, semantic checks on the output of an OCR system can help
detect attacks, but relying on humans to perform this check\textemdash
determine if the OCR output is ``meaningful'' and, if not, compare it
with the input image\textemdash would defeat the purpose of OCR.

We are not aware of any automated system that can check whether the text
produced by OCR ``makes sense.''  The output resulting from our attacks
is not gibberish.  Overall, it reads like normal English text (this is
not surprising, because the attack only modifies a small fraction of the
words), with an occasional awkwardness or inconsistency.  Therefore,
any system for checking the semantic integrity of the text would need
to be very sensitive to individual words appearing out of context.
If such a system existed, we expect that it would be prone to false
positives and vulnerable to adversarial inputs itself.

Furthermore, as Fig.~\ref{fig:changenums} (a) shows, an attack can target
numbers, dates, and other data that does not affect the semantics of
the overall text.  These attacks are difficult to detect using language
processing techniques.

\section{Related Work}
\label{sec:related}


\paragraphbe{Adversarial examples for computer vision.} 
Recent research has shown that deep learning models
are vulnerable to adversarial examples, where a small
change in the input image causes the model to produce a
different output.  Prior work focused mainly on image classification
tasks~\cite{szegedy2013intriguing,goodfellow2014explaining,papernot2016transferability,carlini2017towards},
where the input is a fixed-size image and the output is a class label.
Carlini and Wagner demonstrated an attack that improves on the prior
state of the art in terms of the amount of perturbation and success
rate~\cite{carlini2017towards}.  Their method of generating adversarial
examples is designed for classification problems and cannot be directly
applied to OCR models.

The Houdini approach~\cite{cisse2017houdini} is based on a family of
loss functions that can generate adversarial examples for structured
prediction problems, including semantic segmentation and sequence
labeling.  Houdini is tailored for minimizing the performance of the
model, as opposed to constructing targeted examples, and is not ideal for
targeted attacks against OCR that aim to trick the model into outputting
a specific text chosen by the adversary.

Adversarial examples have been demonstrated for other
computer vision tasks, such as semantic segmentation and
object detection~\cite{xie2017adversarial}, visual question
answering~\cite{xu2017can}, and game playing~\cite{huang2017adversarial}.
These approaches are based on model-specific or task-specific loss
functions for generating adversarial examples and cannot be directly
applied to OCR models.

\paragraphbe{Adversarial examples for NLP.} 
Just like computer vision models are vulnerable to adversarial
perturbations in the image domain, NLP models, too, are vulnerable
to adversarial perturbations in the text domain.  An adversary
can substitute words in the input text to fool many text classification
model~\cite{reddy2016obfuscating,mahler2017breaking,zhao2017generating,papernot2016crafting}.
These approaches require careful word-level substitution, which limits
the performance and power of the attack.

Generating adversarial natural text is harder than generating adversarial
images.  Because of the discrete nature of text, each word carries much
more semantic meaning than each pixel in an image.  It is difficult to
design an attack that modifies words in a way that would not be noticed
by a human.

When NLP models operate on the output of OCR models, the attack surface
is much larger.  The adversary can operate in the image domain and
transform pixels as needed.  He still has to generate targeted attacks
against the OCR model that cause it to output specific character strings,
but in many scenarios he does not need to ensure that these strings are
syntactically or semantically correct as long as they have the desired
effect on the NLP model that consumes them.  This significantly increases
the power of adversarial examples.  Whereas most prior work is concerned
only with the classification error of the NLP model, our attack gives
the attacker full control over the model's predictions and its confidence.

Recent work has also shown that small modifications to words such
as adding random characters and introducing typos can degrade the
performance of models for NLP tasks such as classification and machine
translation~\cite{samanta2017towards,hosseini2017deceiving,belinkov2017synthetic}.
These modifications can be easily integrated with our attack because their
core idea is to produce visually similar words that a human will ignore.

\paragraphbe{Adversarial examples for speech recognition.}
Speech recognition is similar to OCR in the sense that it, too,
aims to assign a sequence of character labels to an input (an
audio recording in the case of speech recognition, an image
in the case of OCR).  Prior work has shown how to generate
mangled, unintelligible, and even inaudible audio inputs that are
nevertheless recognized as commands or speech by speech recognition
systems~\cite{vaidya2015cocaine,carlini16hidden,zhang2017dolphinattack}.
By contrast, we do not aim to generate incomprehensible inputs.
Our goal is to generate images that have the visual appearance
of human-understandable text yet are recognized as a different,
attacker-specified text.

Most recent results on adversarial examples for speech recognition
(developed concurrently with our work) include targeted attacks that
are close to clean audio inputs~\cite{carlini2018audio}.  In this case,
the attacker can set the target to any desired output; in our case, the
targets are limited to the words that are somewhat visually similar to
those in the original clean image.

\begin{figure}
\centering
\fbox{
\includegraphics[width=0.25\textwidth]{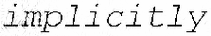}
}
\caption{This adversarial example is rejected by Tesseract even though
its $L_2$ distance to the clean image is only 2.34.}
\label{fig:rejexample}
\end{figure}

A key distinction between the speech recognition models and optical
recognition models is that the former are explicitly designed to work
in noisy environments~\cite{googlespeech,pmlr-v48-amodei16}.  Therefore,
speech-to-text models accept, and attempt to transcribe, sound recordings
with minor squeaks and noises that do not affect human perception
(and as prior work has shown, even sounds that are unintelligible to
humans).  By contrast, Tesseract rejects inputs with relatively minor
perturbations\textemdash see an example in Fig.~\ref{fig:rejexample}.
This greatly limits the space of feasible adversarial examples for
OCR models.

Furthermore, the alphabet for the label sequences produced by the
speech recognition models has size of 26, corresponding to 26 English
characters.  The alphabet for Tesseract's outputs is 110, which includes
upper- and lower-case English characters, numbers, and special symbols.
Larger alphabets make targeted attacks harder.

\section{Conclusions and Future Work}

We demonstrated that OCR systems based on deep learning are vulnerable to
targeted adversarial examples.  Minor modifications to images of printed
text cause the OCR system to ``recognize'' not the word in the image
but its semantic opposite chosen by the adversary.  This enables the
adversary to craft adversarial documents whose meaning changes after
they pass through OCR.  Our attack also has a significant impact on
the NLP applications that use OCR as a preprocessing step, enabling the
adversary to control both their output and their reported confidence.
To the best of our knowledge, this is the first instance of adversarial
examples against sequence labeling models in the image domain.

The adversarial examples in this paper were developed for the latest
version of Tesseract, a popular open-source OCR system based on deep
learning.  They do not transfer to the legacy version of Tesseract, which
employs character-based recognition.  Transferability of adversarial
images across different types of OCR models is an open problem.

Physical realizability, i.e., whether it is possible to create
physical documents whose meaning changes after they are scanned
and processed by OCR, is an interesting topic for future research.
In Section~\ref{sec:physical}, we demonstrated that some of our
adversarial examples are physically realizable.  In general, however,
image perturbations that are sufficiently large to survive the
scanning exceed the amount of noise that Tesseract can tolerate in its
input images.  It remains an open question how to develop adversarial
perturbations for printed natural text that (a) are small enough so they
affect only a single word, (b) do not significantly change the appearance
of this word to a human reader, (c) yet are large enough so they are
preserved when the image is scanned by a commodity scanner, and (d)
cause the OCR system to output a different word chosen by the adversary.

\paragraphbe{Acknowledgements.}
Thanks to Tom Ristenpart for suggesting that OCR systems should be
vulnerable to adversarial examples and to Jasmine Kitahara for experiments
with adversarial images of digits.  This work is partially supported by
a grant from Schmidt Sciences.

\newpage
\clearpage
{\footnotesize \bibliographystyle{acm}
\bibliography{citation}}

\balance


\end{document}